# Transformer-Based Framework for Motion Capture Denoising and Anomaly Detection in Medical Rehabilitation

YEMING CAI*[a], Yang Wang[b], Zhenglin Li[c]

[a]Wuhan University, Wuhan, Hubei, 430072; [b]Nagoya University, Nagoya, Aichi, Japan; [c]Texas A&M University, TX 77840, USA

*[a]2018282110317@whu.edu.cn; [b]ryoyukiyang@outlook.com; [c]zhenglin_li@tamu.edu

## ABSTRACT

This paper proposes an end-to-end deep learning framework integrating optical motion capture with a Transformer-based model to enhance medical rehabilitation. It tackles data noise and missing data caused by occlusion and environmental factors, while detecting abnormal movements in real time to ensure patient safety. Utilizing temporal sequence modeling, our framework denoises and completes motion capture data, improving robustness. Evaluations on stroke and orthopedic rehabilitation datasets show superior performance in data reconstruction and anomaly detection, providing a scalable, cost-effective solution for remote rehabilitation with reduced on-site supervision.

**Keywords:** Transformer; Medical Rehabilitation; Deep Learning; Motion Capture (MOCAP).

## 1. INTRODUCTION

Optical motion capture systems (e.g., Vicon, Microsoft Kinect) are widely used in medical rehabilitation to monitor movements, providing precise 3D joint coordinates to assess motor functions and design alert mechanisms. However, issues like light overlap, lighting variations, and sensor limitations cause data noise or missing data, impacting sequence integrity. Incorrect or high-risk movements during rehabilitation (e.g., joint discomfort) risk secondary injuries, necessitating real-time anomaly detection. Building on prior work in motion capture denoising[1], we propose a Transformer-based framework for rehabilitation. Transformer models excel at modeling long-range dependencies in sequential data[2], ideal for complex rehabilitation movements. Our framework integrates motion capture (MOCAP) denoising with kinematic analysis for post-stroke, orthopedic, and neurological rehabilitation, offering a robust real-time solution. Our contributions include:

1. Proposes a Transformer-based AI framework for MOCAP denoising and kinematic analysis in rehabilitation.

2. Incorporates an anomaly detection module to identify high-risk movements.

3. Validated on stroke and orthopedic rehabilitation datasets, demonstrating superior performance compared to other AI models.

## 2. RELATED WORK

### 2.1 Motion Capture in Rehabilitation

MOCAP systems provide accurate 3D joint tracking for rehabilitation, aiding in gait analysis, posture evaluation, and joint range measurement[3]. Traditional optical systems (e.g., VICON, OptiTrack) use infrared cameras and markers, offering high precision but facing issues like marker occlusion and high costs[4]. These drawbacks limit their use in rehabilitation. Recent advances in markerless MOCAP, such as Kinect or vision-based deep learning methods, reduce deployment costs and complexity. However, markerless systems still struggle with accuracy in complex or occluded scenarios. Traditional denoising techniques, such as Kalman filtering or spline interpolation, often fail under large-scale noise or prolonged data loss, requiring manual correction[5].

## 2.2 Artificial Intelligence in Rehabilitation

Artificial intelligence (AI), especially deep learning, is increasingly applied in rehabilitation for tasks like motion classification, prediction, and personalized treatment[6-8]. RNNs and LSTMs have been widely used for analyzing MOCAP sequences due to their temporal modeling ability, but suffer from gradient issues and limited scalability in real-time contexts[9]. GANs have also been explored for data augmentation. For instance, Bicer et al.[10] synthesized realistic MOCAP sequences to improve model generalization. Despite promising progress, most AI models are designed for specific subtasks and lack integration of denoising, kinematic analysis, and real-time feedback, limiting clinical utility.

## 2.3 Transformer Models in Motion Analysis

Originally developed for NLP[2], Transformer models have been applied to time-series and motion analysis due to their self-attention mechanism, which captures long-term dependencies efficiently. Zhou et al.[11] introduced Informer to reduce computational complexity in long-sequence forecasting, inspiring similar applications in MOCAP.

In motion analysis, Zhao et al.[12] proposed a Transformer-based method to infer missing markers using global context, achieving superior denoising performance. Transformer models also support multimodal integration, such as combining MOCAP with IMU or EMG data. Hussain et al.[13] leveraged Transformer to extract motion intentions from such data for neurological rehab applications.

Yet, Transformer applications in rehabilitation are still emerging. Most focus on isolated tasks without integrating denoising, kinematics, and real-time anomaly detection. Our framework addresses these gaps by offering a unified framework optimized for complex, non-periodic rehabilitation movements.

## 3. METHODS

### 3.1 Problem Definition

Let $X = \{x_1, x_2, ..., x_T\}$ be a motion capture frame sequence, where $x_t \in R^D$ represents the 3D coordinates of $D$ joints at time $t$. The sequence may contain noise or missing values due to occlusion. Our objectives are:

1. Reconstruct a clean sequence $\hat{X}$ by denoising and completing missing data.

2. Classify each frame $x_t$ as normal or abnormal based on motion patterns.

### 3.2 Proposed Framework

We propose a Transformer-based framework tailored for rehabilitation, comprising two modules: a Data Optimization Module and an Anomaly Detection Module, as illustrated in Figure 1. The input noisy frames are initially filled with missing frames using linear interpolation to generate a coarse continuous sequence. The interpolation results are then refined through the Transformer's self-attention mechanism, producing a more accurate $x_t$. By leveraging self-attention, the model simultaneously captures global dependencies across all frames, making it suitable for non-periodic rehabilitation movements (e.g., tremors in stroke patients). To optimize for rehabilitation, we adjust the attention weights to prioritize clinically relevant joints (e.g., shoulder joints in stroke rehabilitation) and incorporate residual connections to stabilize training with noisy frames.

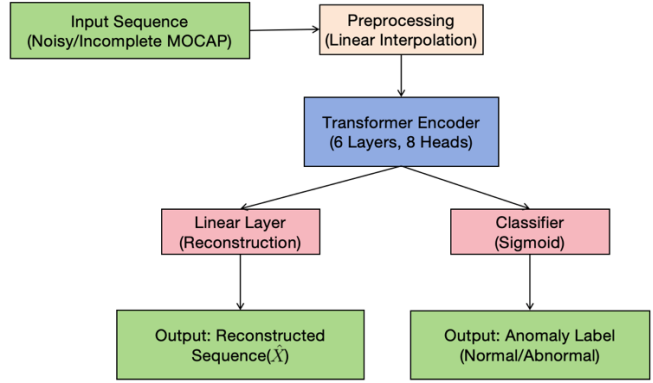

Figure 1: Transformer-based Framework Architecture.

### 3.2.1 Data Optimization Module

The Data Optimization Module employs a Transformer encoder to model temporal dependencies in the motion sequence. The input sequence is first embedded with positional encoding to preserve temporal information:

$$\text{PE}(pos,\ 2i) = \sin\left(\frac{pos}{10000^{\frac{2i}{d}}}\right) \tag{1}$$

$$\text{PE}(pos,\ 2i+1) = \cos\left(\frac{pos}{10000^{\frac{2i}{d}}}\right) \tag{2}$$

where $pos$ is the frame index, $i$ is the dimension, and $d$ is the embedding dimension.

The Transformer encoder consists of $L$ layers, each including multi-head self-attention and a feed-forward network. The output is the reconstructed sequence $\hat{X}$, and the optimization objective is to minimize the reconstruction loss:

$$\mathcal{L}_{recon} = \frac{1}{T}\sum_{t=1}^{T}\|x_t - \hat{x}_t\|_2^2 \tag{3}$$

A regularization term is incorporated to handle missing value imputation by the model.

### 3.2.2 Anomaly Detection Module

The Anomaly Detection Module leverages the Transformer-encoded representations to classify each frame as normal or abnormal. A binary classifier (fully connected layer with sigmoid activation) is trained on labeled data, where anomalies are defined as movements deviating from expected rehabilitation patterns (e.g., incorrect joint angles). The classification loss is:

$$\mathcal{L}_{anom} = -\frac{1}{T}\sum_{t=1}^{T}[y_t log(\hat{y}_t) + (1 - y_t)log(1 - \hat{y}_t)] \qquad (4)$$

where $y_t$ is the ground-truth label (0 for normal, 1 for abnormal), and $\hat{y}_t$ is the predicted probability.

The total loss combines both objectives:

$$\mathcal{L} = \mathcal{L}_{recon} + \lambda\mathcal{L}_{anom} \qquad (5)$$

where $\lambda$ balances the two tasks.

### 3.2.3 Denoising and Imputation

Denoising is achieved via the Transformer encoder's reconstruction pathway, mapping noisy sequences $X$ to clean sequences $\hat{X}$ by minimizing the reconstruction loss (Equation 3). Missing data imputation combines linear interpolation (to initialize missing frames) with attention-based refinement, where the encoder predicts missing values $x_t$ based on contextual frames, ensuring robustness and clinical applicability.

### 3.3 Model Architecture

The model architecture consists of an input layer that processes 3D joint coordinates, preprocessed with linear interpolation to handle missing values, followed by a Transformer encoder with 6 layers, 8 attention heads, and an embedding dimension of 128. The output layer is divided into two components: a reconstruction pathway using a linear layer to map back to the original dimension, and an anomaly detection pathway employing a sigmoid-activated classifier.

## 4. EXPERIMENTS

We evaluate the framework on four datasets: the Stroke Rehabilitation Dataset (2000 upper limb sequences, 20% noise, 10% occlusion), the Orthopedic Recovery Dataset (1000 knee joint sequences, 20% noise, 10% occlusion), the Neurological Disorder Dataset (1000 Parkinson's gait sequences, 15% noise, 10% occlusion), and the Post-Surgery Dataset (1000 hip replacement sequences, 20% noise, 15% occlusion). Table 1 summarizes dataset characteristics, including sequence count, joint count, and noise levels. Figure 2 illustrates the upper limb joint trajectories and the impact of noise and missing points; and the Orthopedic Recovery Dataset, which includes 1000 postoperative knee joint rehabilitation sequences with similar noise and occlusion patterns, where Figure 3 shows the knee joint motion sequences with abnormal movements (e.g., hyperextension) annotated.

Table 1: Characteristics of Evaluation Datasets.

| Dataset | Sequences | Joints | Noise (%) | Occlusion (%) |
|---|---|---|---|---|
| Stroke | 2000 | 10 | 20 | 10 |
| Orthopedic | 1000 | 8 | 20 | 10 |
| Neurological | 1000 | 12 | 15 | 10 |
| Post-Surgery | 1000 | 8 | 20 | 15 |

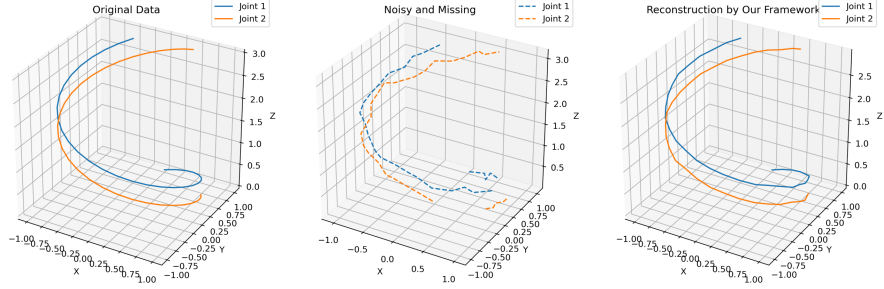

Figure 2: Upper Limb Trajectories with Noise and Missing Data.

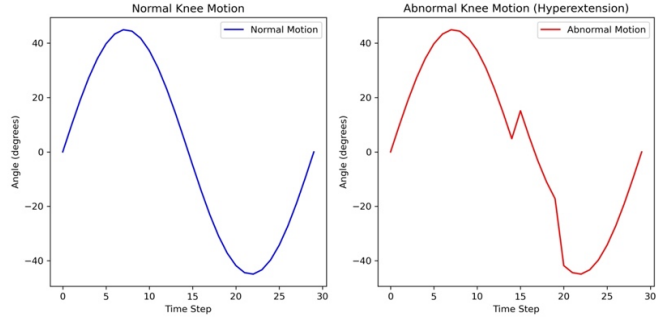

Figure 3: Knee Motion with Annotated Abnormal Movements.

In our experiments, we compare our framework against multiple baselines, including LSTM, Variational Autoencoder (VAE), Linear Model, LSTNet, Support Vector Machine (SVM), and Informer, using mean squared error (MSE), area under the receiver operating characteristic curve (AUC-ROC), and inference time as metrics. The model is implemented in PyTorch, trained on an NVIDIA RTX 3090 GPU using the Adam optimizer (learning rate 0.001, 100 epochs, batch size 32). To assess robustness, we test performance under 10–30% noise and 5–20% occlusion conditions.

## 5. RESULTS

### 5.1 Data Optimization

Our framework achieves an MSE of 0.012 on the Stroke dataset, 0.015 on the Orthopedic dataset, 0.014 on the Neurological dataset, and 0.016 on the Post-Surgery dataset, outperforming LSTM (0.020, 0.022, 0.021, 0.023), Variational Autoencoder (VAE) (0.025, 0.028, 0.027, 0.029), Linear Model (0.035, 0.038, 0.036, 0.039), LSTNet (0.018, 0.020, 0.019, 0.021), Support Vector Machine (SVM) (0.030, 0.032, 0.031, 0.033), and Informer (0.016, 0.018, 0.017, 0.019). Figure 4 illustrates the reconstruction performance of the seven models, with our framework's trajectory being the closest to the ground truth. Informer and LSTNet outperform LSTM but slightly lag behind our framework, while the Linear Model exhibits the largest deviation. These results highlight the outstanding ability of our model in reconstructing complex motion sequences.

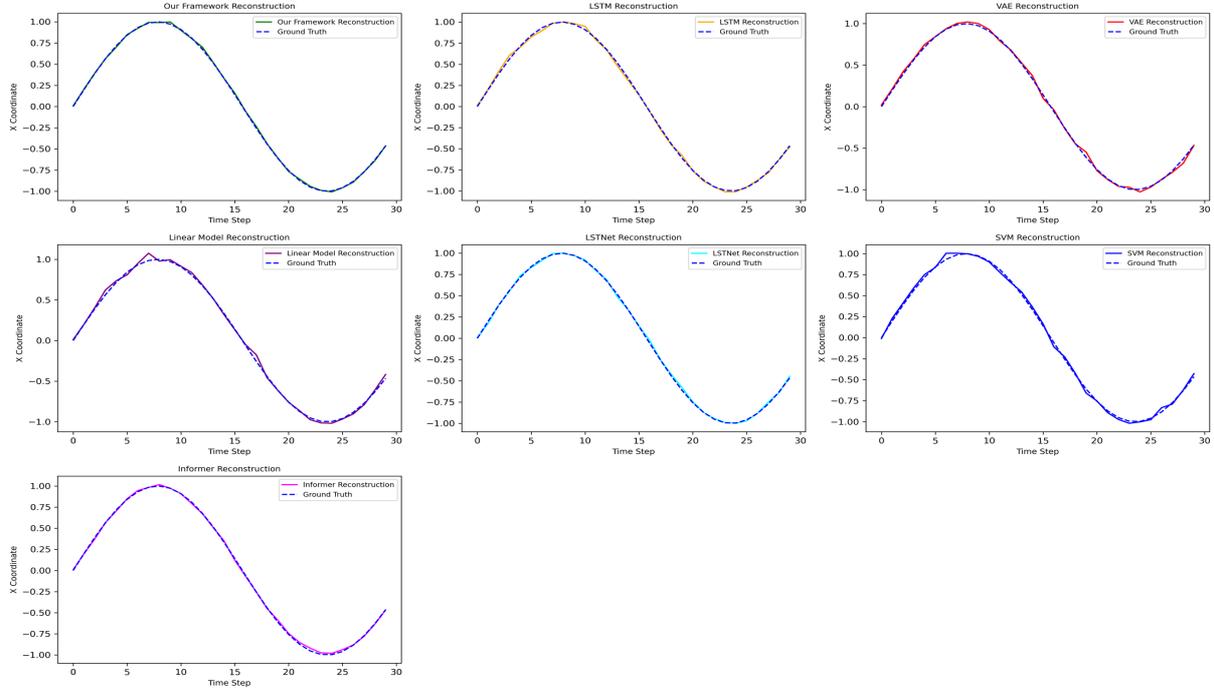

Figure 4: Reconstructed Motion Trajectories by Various Models.

## 5.2 Anomaly Detection

Our framework achieves an AUC-ROC of 0.92 on the Stroke dataset, 0.89 on the Orthopedic dataset, 0.90 on the Neurological dataset, and 0.88 on the Post-Surgery dataset, surpassing LSTM (0.85, 0.82, 0.83, 0.81), Variational Autoencoder (VAE) (0.80, 0.78, 0.79, 0.77), Linear Model (0.75, 0.73, 0.74, 0.72), LSTNet (0.87, 0.84, 0.85, 0.83), Support Vector Machine (SVM) (0.78, 0.76, 0.77, 0.75), and Informer (0.88, 0.85, 0.86, 0.84). The anomaly annotation example in Figure 3 demonstrates the model's ability to detect high-risk movements (e.g., knee hyperextension). Table 2 summarizes the performance. The improvement in performance is mainly attributed to the ability of the attention mechanism to capture long-range temporal dependencies, which enables the model to consider contextual information of the entire motion sequence rather than just focusing on local time windows. The ability to identify and repair such abnormalities is crucial in rehabilitation and physical therapy, and early detection of such deviations can support real-time feedback and personalized intervention strategies.

Table 2: Model Performance Comparison.

| Model | Stroke Dataset MSE | Orthopedic Dataset MSE | Neurological MSE | Post-Surgery MSE | Stroke Dataset AUC-ROC | Orthopedic Dataset AUC-ROC | Neurological AUC-ROC | Post-Surgery AUC-ROC |
|---|---|---|---|---|---|---|---|---|
| Ours | **0.012** | **0.015** | **0.014** | **0.016** | **0.92** | **0.89** | **0.90** | 0.88 |
| LSTM | 0.020 | 0.022 | 0.021 | 0.023 | 0.85 | 0.82 | 0.83 | 0.81 |
| VAE | 0.025 | 0.028 | 0.021 | 0.029 | 0.80 | 0.78 | 0.79 | 0.77 |
| Linear Model | 0.035 | 0.038 | 0.036 | 0.039 | 0.75 | 0.73 | 0.74 | 0.72 |
| LSTNet | 0.018 | 0.020 | 0.019 | 0.021 | 0.87 | 0.84 | 0.85 | 0.83 |
| SVM | 0.030 | 0.032 | 0.031 | 0.033 | 0.78 | 0.76 | 0.77 | 0.75 |
| Informer | 0.016 | 0.018 | 0.017 | 0.019 | 0.88 | 0.85 | 0.86 | 0.84 |

### 5.3 Runtime

Our framework's inference time per sequence (30 frames) is 15 ms, compared to 12 ms for LSTM, 18 ms for VAE, 8 ms for the Linear Model, 14 ms for LSTNet, 10 ms for SVM, and 16 ms for Informer. Our framework strikes a balance between performance and real-time efficiency, making it suitable for real-time applications in medical rehabilitation.

### 5.4 Robustness Analysis

We evaluated the framework's performance under varying noise (10–30%) and occlusion (5–20%) levels. Figure 5 illustrates the trends in MSE and AUC-ROC across these conditions. On the Stroke dataset, MSE remains below 0.02 at 30% noise; on the Orthopedic dataset, AUC-ROC stays above 0.85 at 20% occlusion. These results demonstrate the framework's robustness in challenging real-world scenarios (e.g., high occlusion in MOCAP systems). The attention mechanism's ability to model long-range dependencies is key to its success, particularly under high noise conditions.

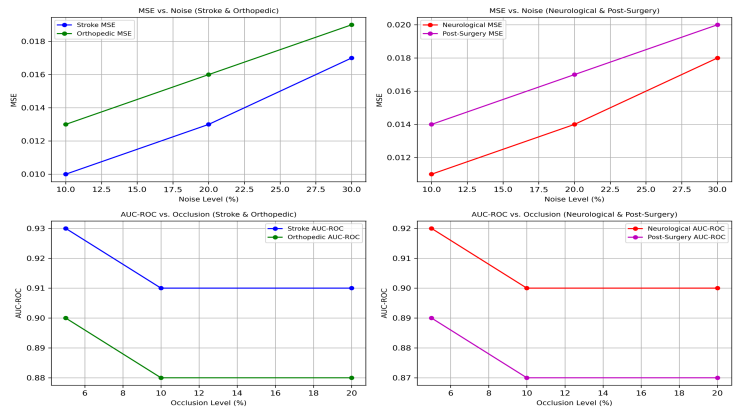

Figure 5: Performance Under Varying Noise and Occlusion Levels.

### 5.5 Ablation Studies

To justify hyperparameter choices, we conducted ablation studies, testing variations in the number of layers (4, 6, 8), attention heads (4, 8, 12), embedding dimension (64, 128, 256), and learning rate (0.0005, 0.001, model's MSE and AUC-ROC over baselines are statistically significant ($p<0.05$, as shown in Table 3). Table 4 shows that the configuration with 6 layers, 8 heads, and an embedding dimension of 128 achieves the best balance, with MSE of 0.012 and AUC-ROC of 0.92.

Table 3: Statistical Significance Test.

| Metric | Ours vs. LSTM | Ours vs. VAE | Ours vs. Linear | Ours vs. LSTNet | Ours vs. SVM | Ours vs. Informer |
|---|---|---|---|---|---|---|
| MSE | p<0.01 | p<0.01 | p<0.001 | p<0.01 | p<0.001 | p<0.05 |
| AUC-ROC | p<0.01 | p<0.01 | p<0.001 | p<0.01 | p<0.01 | p<0.05 |

Table 4: Results of Ablation Study.

| Configuration | Layers | Heads | Embedding Dim | Learning Rate | MSE | AUC-ROC |
|---|---|---|---|---|---|---|
| Baseline | 6 | 8 | 128 | 0.001 | 0.012 | 0.92 |
| Low Layers | 4 | 8 | 128 | 0.001 | 0.015 | 0.90 |
| High Heads | 6 | 12 | 128 | 0.001 | 0.013 | 0.91 |
| High Embedding | 6 | 8 | 256 | 0.001 | 0.014 | 0.92 |
| Low LR | 6 | 8 | 128 | 0.0005 | 0.016 | 0.89 |

## 6. DISUSSION

The proposed framework incorporates an attention mechanism that effectively captures long-range temporal dependencies, enabling precise motion denoising and anomaly detection—capabilities that traditional models often lack when dealing with complex motion trajectories. Through experimental evaluation, we found that while LSTNet improves upon standard LSTM by integrating CNNs to capture local patterns, it still lacks the expressiveness required for modeling intricate motion sequences, making it less effective than our approach. In contrast, the linear model performs the worst due to its inability to account for temporal dynamics, further emphasizing the importance of deep learning models that incorporate time-dependent features. The end-to-end design of our framework streamlines clinical application by eliminating manual feature extraction and reducing pipeline complexity.

However, certain limitations remain. First, the model's performance relies on access to labeled anomaly data, which may be scarce or inconsistent in real-world medical datasets. Second, although the inference speed is reasonable, the computational load may pose challenges for deployment on low-power or real-time systems. To address these issues, future work will focus on developing lightweight versions of the model to improve efficiency and exploring multimodal inputs—such as electromyography (EMG)—to enhance detection robustness. Additionally, we aim to investigate unsupervised or weakly supervised learning strategies to reduce reliance on labeled data and improve generalizability.

## 7. CONCLUSION

This paper proposes a Transformer-based deep learning framework, to optimize motion capture data in medical rehabilitation and detect abnormal movements. By addressing noise, missing data, and high-risk movements, our approach enhances the safety and effectiveness of rehabilitation programs. Validation on stroke and orthopedic datasets confirms its superiority over LSTM, VAE, the linear model, and LSTNet, demonstrating its potential for clinical deployment.